\documentclass[runningheads]{llncs}
\usepackage[T1]{fontenc}
\usepackage{amsmath}
\usepackage{graphicx,verbatim}
\usepackage{hyperref}
\usepackage{color}

\begin{document}
\title{OmniTongue: A Comprehensive MLLM Benchmark for Automated Tongue Diagnosis in Traditional Chinese Medicine}
%
\titlerunning{OmniTongue: A Benchmark for Automated Tongue Diagnosis in TCM}

\author{Liuyi Xu\inst{1} \and
Yun Guo\inst{1} \and
Shusheng Li\inst{2} \and
Xuewen Zhang\inst{1} \and
Yining Qian\inst{2}\thanks{Corresponding author} \and
An-Yang Lu\inst{1}}

\authorrunning{L. Xu et al.}

\institute{College of Information Science and Engineering, Northeastern University, Shenyang 110819, China
\and
College of Computer Science and Engineering, Northeastern University, Shenyang 110189, China\\
\email{2210718@stu.neu.edu.cn}}

\maketitle              
\begin{abstract}
Multimodal Large Language Models (MLLMs) have excelled in Western medical imaging, yet their potential in Traditional Chinese Medicine (TCM) tongue diagnosis remains systematically unevaluated. Tongue diagnosis requires sophisticated global and local perception, serving as a rigorous touchstone for assessing the visual understanding depth of MLLMs. However, existing datasets are limited by single dimensions, coarse granularity, and ambiguous labels, hindering the evaluation of authentic clinical capabilities. To bridge this gap, we introduce OmniTongue, the first comprehensive, fine-grained Visual Question Answering benchmark for TCM tongue diagnosis. This benchmark integrates 1,351 expert-validated images and 21,616 instruction pairs. Its core design features 16 clinical dimensions for multidimensional evaluation, region-specific annotations for pathological localization, and 66 fine-grained labels for precise diagnostic representation. Extensive evaluations of existing MLLMs reveal significant performance gaps, demonstrating that medical-specific models lack domain advantages, while universally poor Macro-F1 scores expose a reliance on statistical shortcuts rather than genuine understanding of tongue pathology. These findings not only highlight the current limitations of MLLMs in interpreting regional medical features but also underscore OmniTongue's significance for developing robust automated tongue diagnosis models. The code and dataset will be made publicly available in a future release.
\end{abstract}

%
%
%

\section{Introduction}

Vision-language models \cite{radford2021clip,liu2023llava,openai2023gpt4} and medical adaptations \cite{tu2024medpalm,li2023llavamed,moor2023medflamingo,zhang2023biomedclip,wu2025radfm,saab2024medgemini} have advanced medical image understanding. However, as a cornerstone of Traditional Chinese Medicine (TCM), tongue diagnosis \cite{wang2020artificial,li2017tonguematting,yang2025tonguenet,chen2025drtongue,zhang2023tonguecoating} has not been systematically evaluated. Tongue diagnosis requires sophisticated global perception and region-conditioned attribute recognition to interpret systemic health from the color, morphology, and distribution of the tongue body and coating. This demand for capturing global features while precisely localizing pathological signs in specific regions (e.g., tip, middle, or root) makes tongue diagnosis an ideal touchstone for assessing the visual understanding depth of MLLMs, extending beyond the scope of general-domain benchmarks \cite{liu2024mmbench,yue2024mmmu}.

Existing tongue diagnosis datasets fall short of supporting such evaluations due to three critical limitations. First, dimensional scarcity restricts most datasets to single global labels \cite{liang2025ifrcnet,li2019toothmarked}, precluding the assessment of comprehensive diagnostic capabilities across multiple clinical dimensions. Second, regional tongue-sign labels are scarce, despite progress in general medical image segmentation such as MedSAM \cite{ma2024medsam}. Third, coarse labeling merges diagnostically distinct features into generic categories, which obscures critical pathological information essential for precise tongue phenotype recognition. This lack of region-level supervisory signals renders existing datasets ineffective for evaluating the true capabilities of MLLMs.

To bridge this gap, we introduce OmniTongue, the first comprehensive and fine-grained Visual Question Answering (VQA) benchmark specifically designed for TCM tongue diagnosis. Validated by senior experts, the benchmark comprises 1,351 high-definition images and 21,616 multiple-choice items. OmniTongue is distinguished by three core characteristics: multidimensional annotation of 16 clinical dimensions for comprehensive evaluation; region-level supervision for the tip, middle, and root to evaluate localization; and fine-grained classification involving 66 diagnostic labels, such as the distinction between light red, crimson, and dark purple, to enforce the learning of precise pathological representations. 

Our systematic evaluation of nine representative MLLMs on OmniTongue reveals significant challenges. First, medical-specific models fail to exhibit anticipated domain advantages in region-conditioned diagnostic tasks, reflecting a generalization bottleneck in specialized tongue diagnostics. Second, high accuracy alongside poor Macro-F1 suggests possible majority-class shortcuts. General-purpose grounding methods \cite{you2024ferret,peng2023kosmos2,chen2023shikra} address spatial reference but not tongue-specific diagnosis. This finding not only exposes fundamental limitations of current models but also underscores the difficulty and utility of OmniTongue as an evaluation benchmark. 

Our main contributions are summarized as follows:
\begin{enumerate}
    \item We introduce OmniTongue, the first comprehensive and fine-grained VQA benchmark for TCM tongue diagnosis, establishing a systematic standard for assessing MLLMs via 16 diagnostic dimensions, 66 fine-grained labels, and region-level annotations.
    \item We conduct an in-depth evaluation of nine mainstream MLLMs, providing the most extensive benchmark assessment in the tongue diagnosis domain to date.
    \item We expose the fundamental limitations in regional diagnostic attribute recognition and the misleading nature of conventional metrics, guiding the development of anatomy-aware tongue diagnosis models.
\end{enumerate}

\section{OmniTongue: A Comprehensive Benchmark for Tongue Diagnosis}

\begin{figure}[!ht] 
  \centering
  \includegraphics[width=1\textwidth]{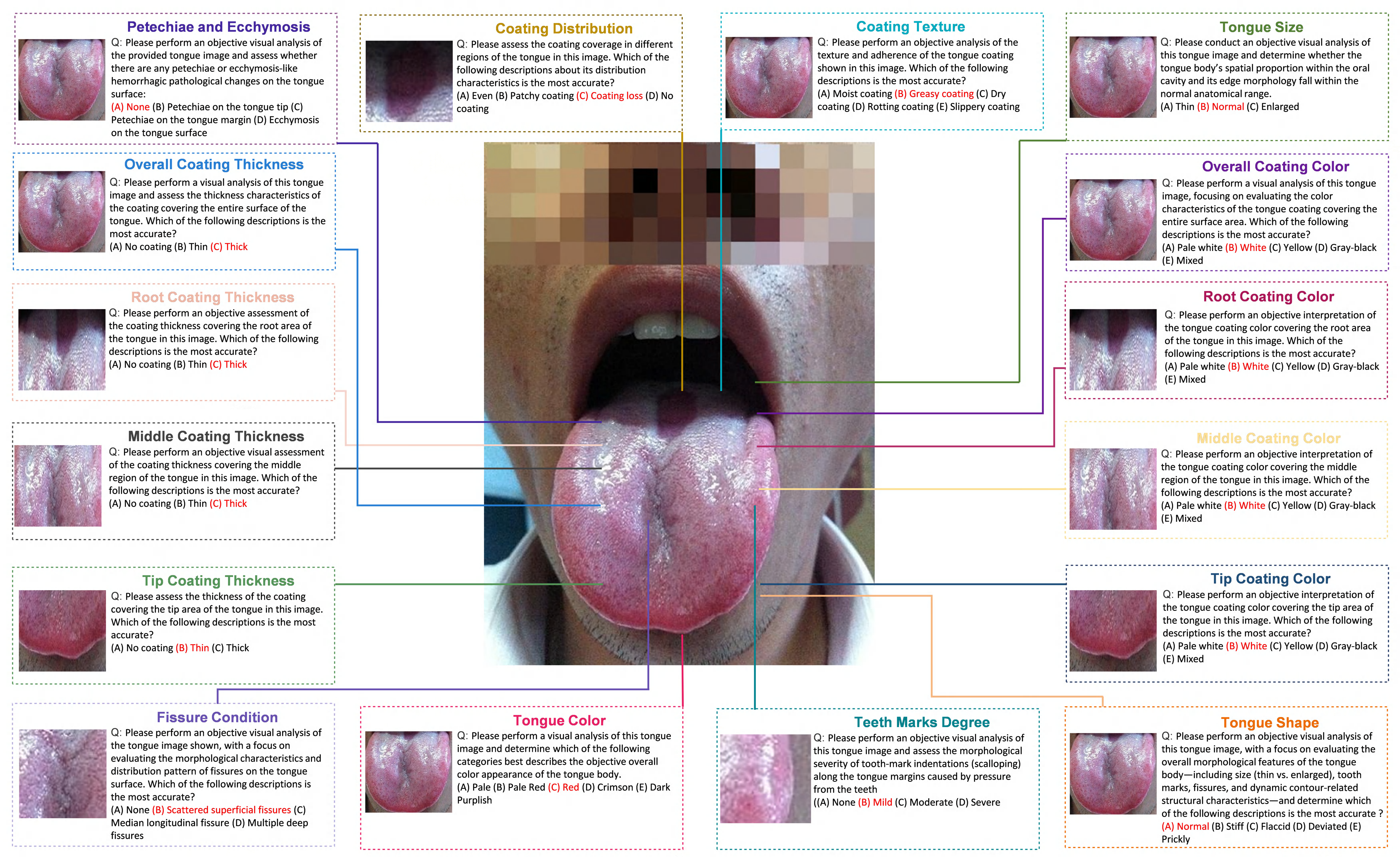}
  \caption{\textbf{Overview of our OmniTongue dataset.} OmniTongue covers the majority of critical diagnostic features and regional characteristics of the human tongue, such as tongue size, shape, color, fissure conditions, teeth marks, as well as coating attributes (thickness, color, distribution, and texture) across the tip, middle, and root areas.} 
  
  \label{fig:framework}
\end{figure}

To facilitate anatomy-aware medical AI, we introduce OmniTongue (Fig.~\ref{fig:framework}), a large-scale benchmark for tongue diagnosis featuring dense regional annotations.

\subsection{Data Collection and Annotation}
The dataset comprises 1,351 high-resolution tongue images collected 
from three clinical sites using calibrated digital cameras under 
standardized lighting over a multi-month period. Each image was 
independently annotated by two senior TCM physicians with over 10 
years of clinical experience across all 16 dimensions, and a third 
senior expert adjudicated disagreements to produce the final 
consensus label. The resulting inter-annotator Cohen's $\kappa$ 
reaches 0.709 per dimension and 0.820 when pooled, with 88.4\% 
raw agreement. Age and sex metadata were not collected in the 
current release and will be addressed in future work. 
Unlike previous datasets that focus on single-label classification, OmniTongue provides dense supervision across a comprehensive set of 16 clinical dimensions. Each image is associated with a structured feature vector that systematically captures these 16 dimensions across clinical groupings: \textbf{Global Tongue Body} (6 dimensions: color, size, shape, teeth marks, cracks, and petechiae); \textbf{Global Tongue Coating} (4 dimensions: overall color, thickness, distribution, and texture); and \textbf{Regional Tongue Coating} (6 dimensions: color and thickness specifically localized to the tip, middle, and root, which anatomically map to distinct internal organ systems). Following standard TCM practice, the Tip, Middle, and Root denote 
the anterior, central, and posterior thirds of the tongue along its 
longitudinal axis, corresponding to the Heart--Lung, 
Spleen--Stomach, and Kidney systems. The detected tongue region is 
split into three equal longitudinal segments during preprocessing.

\begin{figure}[!ht] 
  \centering
  \includegraphics[width=1\textwidth]{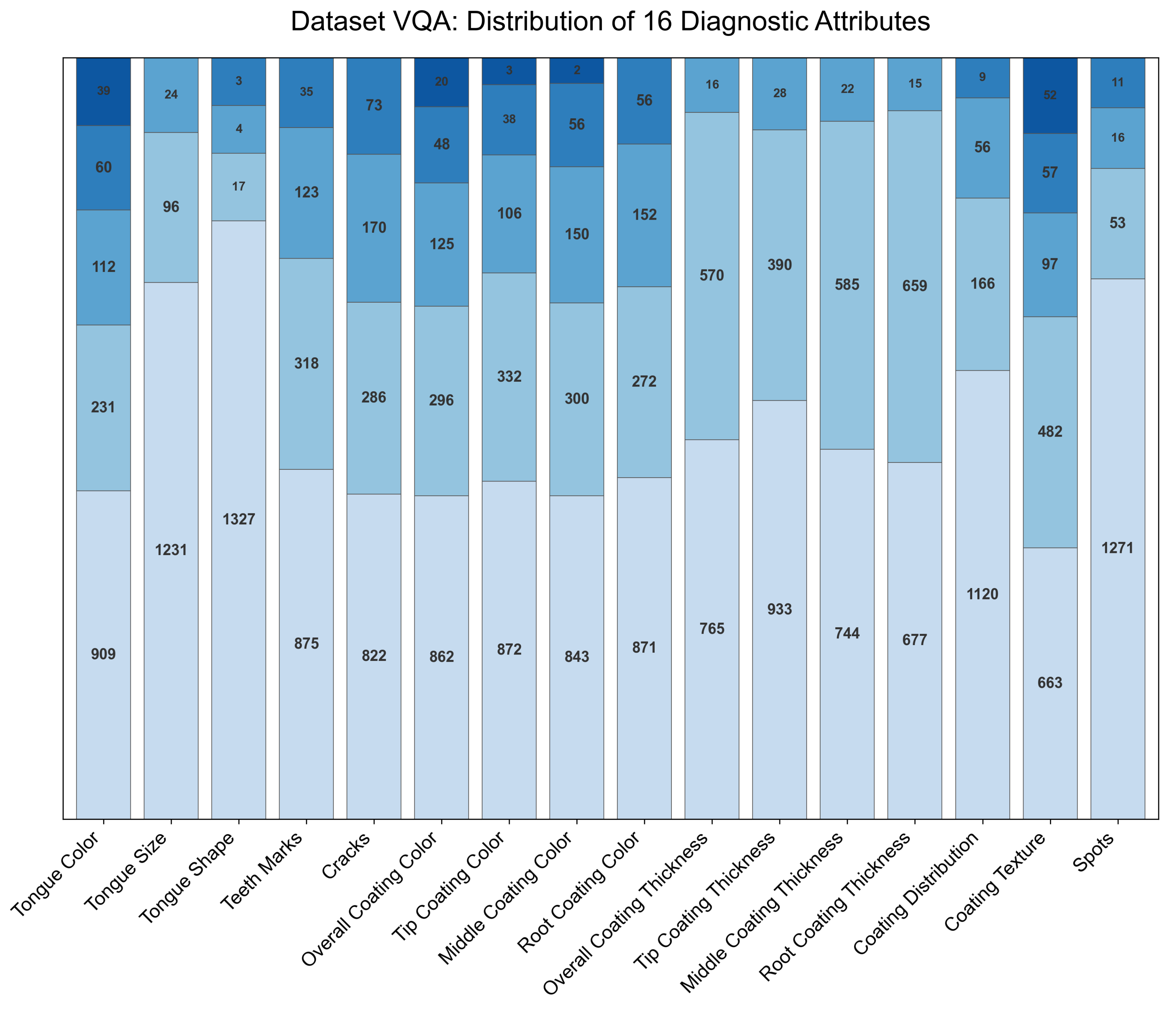}
    \caption{\textbf{Dataset VQA: Distribution of 16 comprehensive clinical dimensions.} Colored blocks within each column denote distinct categories, stacked bottom-up by decreasing sample frequency. OmniTongue exhibits a pronounced long-tailed distribution mirroring authentic clinical scenarios, where common conditions dominate over critical but rare pathological signs. This natural imbalance challenges MLLMs to recognize minority-class local lesions instead of exploiting global statistical shortcuts.}
    \label{fig:taxonomy}
\end{figure}

\subsection{Statistics and Characteristics of the Dataset}
As shown in Figure \ref{fig:taxonomy}, OmniTongue exhibits a long-tailed distribution characteristic of real-world clinical data. For instance, common conditions such as Pale Red Tongue or Thin White Coating appear frequently, whereas severe pathological signs such as Peeled Coating are rare but critical for accurate diagnosis. This severe imbalance presents a significant challenge for existing MLLMs, testing their ability to reliably generalize to critical minority classes.

\subsection{Task Formalization and Benchmark Construction}
Beyond biomedical VQA resources such as PMC-VQA \cite{zhang2023pmcvqa}, OmniTongue targets region-conditioned tongue attributes. By formalizing the transformation of localized clinical annotations into a multimodal Vision-Language Question Answering (VQA) format, we establish a robust proxy metric for assessing models' spatial grounding capabilities.

\textbf{Mathematical Formulation.} Let $\mathcal{I} = \{I_i\}_{i=1}^{N}$ 
denote the set of $N{=}1{,}351$ tongue images. We define $K{=}16$ clinical 
dimensions $\mathcal{D} = \{d^{(k)}\}_{k=1}^{K}$, each with a discrete 
label space $\mathcal{V}^{(k)}$. For each image $I_i$, the expert 
consensus yields a multi-dimensional label vector 
$\mathbf{y}_i = [y_i^{(1)}, \dots, y_i^{(K)}]$ where 
$y_i^{(k)} \in \mathcal{V}^{(k)}$.

\textbf{VQA Generation.} Each label $y_i^{(k)}$ is programmatically 
transformed into a question--answer pair $(q_i^{(k)}, v_i^{(k)})$ via 
template-based query generation and answer formatting, producing the 
benchmark:
\begin{equation}
    \mathcal{D}_{VQA} = \Big\{ \big( I_i, q_i^{(k)}, v_i^{(k)} \big) 
    \;\Big|\; 1 \le i \le N, \; 1 \le k \le K \Big\}
\end{equation}
To prevent models from exploiting lexical patterns, we employ Large 
Language Models to paraphrase generated questions into diverse natural 
language expressions while preserving medical semantics.

This expands the 1,351 images into $|\mathcal{D}_{VQA}| = N \times K 
= 21{,}616$ vision-language instruction pairs. To prevent data leakage, 
all splits are defined at the image level so that VQA items from the 
same source image never cross partitions. The benchmark supports both 
zero-shot evaluation and supervised fine-tuning.

\section{Experiments}
\subsection{Experimental Setup and Evaluation Protocol}

To systematically assess the visual understanding depth of MLLMs in comprehensive tongue diagnosis, we evaluate a representative suite of state-of-the-art models with diverse architectures. To specifically investigate the generalization bottleneck and the efficacy of current paradigms, we categorize our baselines into four groups: (1) \textbf{General-purpose and Scaled Models} (Qwen2.5-VL-7B \cite{qwen2025qwen25vl}, LLaVA-NeXT-7B \cite{liu2024llavanext}, Qwen3-VL-8B \cite{bai2025qwen3vl}, and the InternVL series \cite{chen2024internvl,zhu2025internvl3}) to test if parameter scaling resolves spatial blind spots; (2) \textbf{Efficient Models} (MiniCPM-V series \cite{yao2024minicpmv}) to evaluate resource-constrained performance; (3) A \textbf{Medical-specific Model} (HuatuoGPT-Vision \cite{chen2024huatuogptvisioninjectingmedicalvisual}) to investigate anticipated domain advantages; and (4) A \textbf{Traditional Baseline} (BLIP-2 \cite{li2023blip2}).

We establish a rigorous evaluation protocol across all $K=16$ dimensions by primarily reporting the \textbf{Macro-F1 score} with Top-1 Accuracy tracked as a secondary reference. To provide a theoretical zero-shot lower bound, the expected random guess accuracy for dimension $k$ is defined as $1/|\mathcal{V}^{(k)}|$. The reported `Avg.' denotes the unweighted macro-average across respective dimensions. We strictly prioritize Macro-F1 over standard accuracy because real-world clinical data inherently exhibits a severe long-tailed distribution; relying solely on Top-1 Accuracy over-rewards majority-class predictions and creates a capability illusion, masking models' failures on rare but clinically critical localized signs.

\subsection{Main Results: The Generalization Bottleneck in Tongue Diagnostics}
\begin{table*}[t]
\centering
\fontsize{8}{9}\selectfont
\setlength{\tabcolsep}{1.5pt}

\caption{\textbf{Global Perception Assessment (I):} Benchmark results on OmniTongue for \textbf{Global Tongue Body Features} (Macro-F1 / Acc \%). This evaluates the model's ability to capture holistic morphological and chromatic attributes.}\label{tab:global_body_results}
\begin{tabular}{l|cccccc|c}
\hline
\textbf{Model} & \begin{tabular}[c]{@{}c@{}}\textbf{Body}\\ \textbf{Color}\end{tabular} & \begin{tabular}[c]{@{}c@{}}\textbf{Body}\\ \textbf{Size}\end{tabular} & \begin{tabular}[c]{@{}c@{}}\textbf{Body}\\ \textbf{Shape}\end{tabular} & \begin{tabular}[c]{@{}c@{}}\textbf{Body}\\ \textbf{Crack}\end{tabular} & \begin{tabular}[c]{@{}c@{}}\textbf{Body}\\ \textbf{Mark}\end{tabular} & \begin{tabular}[c]{@{}c@{}}\textbf{Body}\\ \textbf{Spot}\end{tabular} & \textbf{Avg.} \\ \hline
\textit{Random} & \textit{20.0/20.0} & \textit{33.3/33.3} & \textit{25.0/25.0} & \textit{25.0/25.0} & \textit{25.0/25.0} & \textit{25.0/25.0} & \textit{25.6/25.6} \\ \hline
BLIP-2 & 3.8/5.0 & 11.3/18.3 & 19.8/\textbf{98.1} & 22.6/\underline{58.1} & 22.5/60.6 & 28.1/\textbf{94.0} & 18.0/\underline{55.7} \\
LLaVA-NeXT & 16.1/28.7 & 27.5/\underline{59.0} & 25.2/71.8 & 32.0/26.0 & 22.5/34.7 & 25.5/76.0 & 24.8/49.4 \\
MCPM-V2.6 & 17.1/58.3 & \underline{30.4}/48.8 & 14.7/12.3 & 33.0/29.6 & 22.2/26.9 & 15.5/29.7 & 22.2/34.3 \\
MCPM-V4.5 & 24.4/50.4 & 10.4/13.9 & 25.0/\underline{89.6} & 22.4/30.9 & \underline{37.7}/\underline{61.0} & 27.7/71.8 & 24.6/52.9 \\
InternVL2 & 20.2/28.8 & 19.6/27.4 & \underline{30.3}/50.5 & 25.1/21.0 & 13.3/14.6 & 23.4/39.9 & 22.0/30.4 \\
InternVL3 & 12.3/18.5 & 7.1/7.7 & \textbf{31.8}/65.7 & 32.0/44.9 & 23.5/32.3 & \textbf{31.4}/83.0 & 23.0/42.0 \\
Qwen2.5-VL & \textbf{32.6}/\underline{67.0} & 10.2/15.2 & 29.2/57.0 & \underline{36.1}/33.7 & 27.5/32.0 & 26.2/78.2 & \underline{27.0}/47.2 \\
Qwen3-VL & 30.5/65.7 & \textbf{38.7}/\textbf{80.7} & 19.1/53.1 & \textbf{53.8}/\textbf{62.9} & \textbf{40.0}/\textbf{71.9} & \underline{29.0}/\underline{90.3} & \textbf{35.2}/\textbf{70.8} \\
HuatuoGPT & \underline{32.1}/\textbf{68.5} & 5.0/7.9 & 20.9/49.6 & 17.2/20.4 & 13.2/12.1 & 25.5/71.1 & 19.0/38.3 \\ \hline
\end{tabular}

\caption{\textbf{Global Perception Assessment (II):} Benchmark results on OmniTongue for \textbf{Global Tongue Coating Features} (Macro-F1 / Acc \%). These dimensions represent the systemic physiological state through the overall coating presentation.}
\label{tab:global_coating_results}
\begin{tabular}{l|cccc|c}
\hline
\textbf{Model} & \begin{tabular}[c]{@{}c@{}}\textbf{Coating}\\ \textbf{Color}\end{tabular} & \begin{tabular}[c]{@{}c@{}}\textbf{Coating}\\ \textbf{Thick}\end{tabular} & \begin{tabular}[c]{@{}c@{}}\textbf{Coating}\\ \textbf{Texture}\end{tabular} & \begin{tabular}[c]{@{}c@{}}\textbf{Coating}\\ \textbf{Distr}\end{tabular} & \textbf{Avg.} \\ \hline
\textit{Random} & \textit{20.0/20.0} & \textit{33.3/33.3} & \textit{20.0/20.0} & \textit{25.0/25.0} & \textit{24.6/24.6} \\ \hline
BLIP-2 & 8.4/21.5 & 20.9/39.5 & 14.5/41.7 & 23.1/\textbf{81.9} & 16.7/46.1 \\
LLaVA-NeXT & 16.2/22.9 & 28.5/37.2 & 18.1/38.5 & 12.2/22.9 & 18.8/30.4 \\
MCPM-V2.6 & 26.2/33.8 & 36.5/60.8 & 38.2/\underline{72.3} & 11.9/15.0 & 28.2/45.5 \\
MCPM-V4.5 & \textbf{31.6}/44.6 & 37.0/49.1 & 33.4/64.9 & 28.4/62.5 & 32.6/55.3 \\
InternVL2 & 23.3/33.2 & 35.4/48.3 & 37.0/60.3 & 29.0/60.5 & 31.2/50.6 \\
InternVL3 & 27.4/38.9 & 23.6/26.9 & 21.2/44.6 & 29.5/\underline{68.3} & 25.4/44.7 \\
Qwen2.5-VL & 27.2/\underline{58.7} & 41.9/59.0 & \underline{46.2}/69.9 & 19.6/22.8 & 33.7/52.6 \\
Qwen3-VL & \underline{30.1}/\textbf{59.5} & \underline{48.0}/71.2 & \textbf{47.0}/\textbf{82.9} & \underline{31.6}/43.5 & \underline{39.2}/\textbf{64.3} \\
HuatuoGPT & 28.5/48.9 & \textbf{48.5}/\textbf{73.4} & 45.9/67.5 & \textbf{38.3}/50.3 & \textbf{40.3}/\underline{60.0} \\ \hline
\end{tabular}
\end{table*}

\begin{table*}[t]
\centering
\fontsize{8}{9}\selectfont
\setlength{\tabcolsep}{2pt} 

\caption{\textbf{Regional Perception Assessment:} Benchmark results on OmniTongue for \textbf{Comprehensive Regional Coating Features} (Macro-F1 / Acc \%). This evaluates models' spatial grounding and diagnostic precision across the Root (R: Kidney), Middle (M: Spleen/Stomach), and Tip (T: Heart/Lung) anatomical zones.}
\label{tab:local_results}

\begin{tabular}{ l | ccc | ccc | c }
\hline
\textbf{Model} & \multicolumn{3}{c|}{\textbf{Coating Color}} & \multicolumn{3}{c|}{\textbf{Coating Thick}} & \textbf{Avg.} \\ 
 & \textbf{Root} & \textbf{Mid} & \textbf{Tip} & \textbf{Root} & \textbf{Mid} & \textbf{Tip} & \\ \hline
\textit{Random} & \textit{25.0/25.0} & \textit{20.0/20.0} & \textit{20.0/20.0} & \textit{33.3/33.3} & \textit{33.3/33.3} & \textit{33.3/33.3} & \textit{27.5/27.5} \\ \hline
BLIP-2 & 10.4/20.2 & 7.9/22.0 & 7.9/24.3 & 16.3/26.9 & 15.4/24.4 & 20.7/40.7 & 13.1/26.4 \\
LLaVA-NXT & 16.5/21.6 & 19.0/24.6 & 17.7/25.2 & 28.8/33.8 & 23.8/26.0 & 32.2/42.0 & 23.0/28.9 \\
MCPM-V2.6 & 24.6/57.1 & 28.1/52.8 & 24.2/49.1 & \underline{37.9}/\underline{60.0} & 39.4/63.4 & 38.9/\underline{72.0} & 32.2/59.1 \\
MCPM-V4.5 & \underline{30.4}/47.3 & \underline{34.8}/53.8 & \underline{27.6}/44.7 & 33.7/46.2 & 36.2/47.9 & 33.4/43.6 & 32.7/47.2 \\
InternVL2 & 13.9/21.1 & 20.5/30.7 & 21.2/33.1 & 22.8/44.8 & 21.0/40.9 & 20.2/30.5 & 19.9/33.5 \\
InternVL3 & 24.5/39.1 & 28.6/47.0 & 23.2/38.6 & 28.9/33.9 & 33.0/40.2 & 29.1/38.3 & 27.9/39.5 \\
Qwen2.5-VL & 25.5/\underline{58.5} & 29.2/\underline{62.4} & 23.3/\underline{57.9} & 33.3/41.7 & 43.9/65.6 & 37.0/61.4 & 32.0/57.9 \\
Qwen3-VL & \textbf{39.6}/\textbf{63.2} & \textbf{36.6}/\textbf{65.2} & \textbf{29.8}/\textbf{62.1} & 32.5/55.7 & \textbf{48.2}/\underline{71.7} & \underline{42.1}/62.8 & \textbf{38.1}/\underline{63.4} \\
HuatuoGPT & 25.5/54.5 & 29.3/60.5 & 23.7/56.1 & \textbf{45.9}/\textbf{68.5} & \underline{47.3}/\textbf{71.7} & \textbf{45.4}/\textbf{74.8} & \underline{36.2}/\textbf{64.3} \\
\hline
\end{tabular}
\end{table*}

In this section, we present a comprehensive evaluation of the selected MLLMs on the OmniTongue benchmark. Rather than relying solely on superficial global metrics, our analysis dissects the models' perceptual capabilities across fine-grained labels and region-specific annotations. The evaluation exposes a severe capability gap, revealing that current models heavily depend on statistical shortcuts rather than genuine biomedical visual understanding. We synthesize these vulnerabilities into three critical findings. 
Tables~\ref{tab:global_body_results}--\ref{tab:global_coating_results} 
support Findings 1--2 and Table~\ref{tab:local_results} provides 
direct evidence for Finding 3.

\textbf{Current Paradigms Fail to Resolve Spatial Blind Spots 
in Tongue Diagnosis.}
Our evaluation reveals that neither general-purpose models nor medical-specific models overcome the localization bottleneck in tongue diagnostics. Among general-purpose models, even the strongest performer Qwen3-VL still exhibits substantial weakness on regional dimensions: for example, InternVL3 achieves only 29.1\% Macro-F1 on Tip Coating Thickness. We note that 
cross-version differences between similarly sized models such as InternVL3 vs.\ InternVL2 and Qwen3-VL vs.\ Qwen2.5-VL may stem from architecture, training data, or alignment rather than scale alone. On the medical side, HuatuoGPT-Vision, despite its domain-specific alignment, remains weak on Tip Coating Color at only 23.7\% Macro-F1, indicating that medical pre-training without explicit spatial grounding does not confer the anticipated advantage. These results demonstrate that current paradigms, whether through scaling or domain specialization, are insufficient to bridge the capability gap in region-level tongue diagnosis.

\textbf{The Capability Illusion: Standard Metrics Mask Reliance on Statistical Shortcuts.}
A deeper analysis of these generalization bottlenecks reveals a critical capability illusion: strong Top-1 Accuracy on prominent global dimensions can coexist with substantially lower Macro-F1 scores on long-tailed regional dimensions. For instance, Qwen3-VL achieves high accuracy on global dimensions such as Tongue size (80.7\%) and Spots (90.3\%), yet its Macro-F1 on Spots collapses to 29.0\%. 
Given that clinically critical localized signs naturally constitute the tail of the real-world clinical distribution, this discrepancy highlights that current MLLMs fail to deliver consistent performance across the comprehensive set of 66 fine-grained labels across all 16 dimensions.
Here, ``statistical shortcuts'' refer to predictions driven by 
the marginal class prior $P(y)$ rather than region-conditioned 
evidence $P(y \mid \text{image}, \text{region})$, as revealed by 
the large Top-1 Accuracy versus Macro-F1 gap.
Instead of reflecting robust phenotype recognition, the strong accuracy likely arises from statistical correlations dominated by majority classes rather than genuine biomedical understanding.

\textbf{The Localization Bottleneck: Lack of Explicit Anatomical Grounding.}
The fundamental root cause of this capability illusion lies in the models' inability to perform precise pathological localization. Table~\ref{tab:local_results} explicitly demonstrates this by isolating regional coating dimensions to evaluate the models' ability to differentiate between the tongue's Root, Middle, and Tip. The results reveal substantial accuracy variance and inconsistent predictions across these adjacent regions. For example, LLaVA-NeXT's coating thickness Macro-F1 decreases significantly from 32.2\% at the Tip to 23.8\% at the Middle. Such variance suggests that standard visual-language alignment does not consistently anchor region-specific concepts to the intended anatomical areas. These results suggest possible misattribution of visual evidence across regions; related object-hallucination behavior has been studied in prior work \cite{li2023pope}. These findings reinforce the core motivation behind OmniTongue: region-conditioned medical diagnosis fundamentally requires explicit anatomical grounding and high-density expert supervision across 16 anatomical dimensions.

\section{Conclusion}
In this paper, we introduced OmniTongue, the first comprehensively annotated benchmark evaluating fine-grained, region-level perception in Automated Tongue Diagnosis. By providing high-density expert supervision across 16 anatomical dimensions, it addresses the critical lack of spatial granularity in existing tongue datasets. Our systematic evaluation of state-of-the-art MLLMs exposes a severe capability illusion. While exhibiting robust global proficiency, MLLMs rely heavily on statistical shortcuts, consistently failing at precise region-specific pathological localization. Our findings underscore that merely scaling parameters or utilizing medical-specific models is insufficient; explicit anatomical grounding is essential. 
Our current evaluation focuses on nine representative open-source 
models; we plan to extend coverage to closed-source and additional 
medical-specific MLLMs in future work.
OmniTongue establishes a rigorous standard for developing interpretable, anatomy-aware tongue diagnosis MLLMs.

\subsubsection{\discintname}
The authors have no competing interests to declare that are relevant to the content of this article.

%
%
%
\bibliographystyle{splncs04}
\bibliography{refs}

\end{document}